\documentclass{styles/svproc}
\usepackage{url}
\usepackage{graphicx}   
\usepackage{rotating}   
\usepackage{booktabs}   

\begin{document}
\mainmatter              
\title{Tenyidie Syllabification corpus creation and deep learning applications}
\titlerunning{Tenyidie Syllabification corpus creation and deep learning applications}  
%
\author{Teisovi Angami*\inst{1} \and Kevisino Khate\inst{1}}

\authorrunning{Teisovi Angami et al.} 
%
\tocauthor{Teisovi Angami and Kevisino Khate}
\institute{Department of Information Technology, Nagaland University\\
\email{*teisovi@nagalanduniversity.ac.in, kevisinokhate@gmail.com}}

\maketitle              

\begin{abstract}
The Tenyidie language is a low-resource language of the Tibeto-Burman family spoken by the Tenyimia Community of Nagaland in the north-eastern part of India and is considered a major language in Nagaland. It is tonal, Subject-Object-Verb, and highly agglutinative in nature. Being a low-resource language, very limited research on Natural Language Processing (NLP) has been conducted. To the best of our knowledge, no work on syllabification has been reported for this language. Among the many NLP tasks, syllabification or syllabication is an important task in which the given word syllables are identified. The contribution of this work is the creation of 10,120 syllabified Tenyidie words and the application of the Deep Learning techniques on the created corpus. In this paper, we have applied LSTM, BLSTM, BLSTM+CRF, and Encoder-decoder deep learning architectures on our created dataset. In our dataset split of 80:10:10 (train:validation:test) set, we achieved the highest accuracy of 99.21\% with BLSTM model on the test set. This work will find its application in numerous other NLP applications, such as morphological analysis, part-of-speech tagging, machine translation, etc, for the Tenyidie Language.
\keywords{Tenyidie; NLP; syllabification; deep learning; LSTM; BLSTM; CRF; Encoder-decoder}
\end{abstract}

\section{Introduction}
The Tenyidie language is a low-resource language of the Tibeto-Burman~\cite{1} Language family spoken by the Tenyimia Community and is considered a major language of Nagaland in the north-eastern part of India. In its linguistic characteristics, it is a tonal~\cite{2,3,4,5,6} (posits four/five tones), Subject-Object-Verb~\cite{4,7,3} and highly agglutinative~\cite{4} language possessing high inflected forms of the root word. The orthography of the Tenyidie language uses the Roman (Latin) alphabet and uses \textit{space} as a word divider or word boundary, thereby eliminating the problem of word segmentation, which is a challenging task in many languages. There are numerous tasks in natural language processing (NLP), and a few of them are tasks such as: \textit{Morphological segmentation, Part-of-speech (POS) tagging, Parsing, syllabification, Word segmentation, Machine translation, Named entity recognition (NER), Relationship extraction, Sentiment analysis, Word sense disambiguation, etc.}. Among these, syllabification is an important task in which the given word syllables are identified. This syllable information is then used in many other NLP tasks, such as understanding morphological characteristics of a language, part-of-speech tagging, named-entity recognition, etc.

The main aim of our work is the creation of 10,120 words syllabified Tenyidie words and the application of the Deep Learning techniques on the created corpus. To the best of our knowledge, no work on syllabification has been reported for the Tenyidie language. In this paper, we have applied LSTM, BLSTM, BLSTM+CRF, and Encoder-decoder deep learning architectures on our created dataset. 

The rest of the paper is organized as follows: Section~\ref{section: The Tenyidie Language} gives an introduction to the Tenyidie Language, Section~\ref{section:related works} gives an overview of the related works, Section~\ref{section: corpus_generation} gives a description of the syllabified corpus generation and provides a statistics of the Tenyidie syllables in our annotated corpus, Section~\ref{section: learning_model} explains the learning models experimental setup, Section~\ref{section: experimental_results} reports and discusses the experimental results, and Section~\ref{section: conclusion} draws the conclusion and discusses future works.

\section{The Tenyidie Language}
\label{section: The Tenyidie Language}
This section provides a brief overview of the character set and syllable structure of the Tenyidie Language.

\subsection{Character Set for Tenyidie Language} 

The Tenyidie character set is similar to the English Alphabet except that it doesn’t have the letters ‘Q’ \& ‘X’ and includes a special letter ‘Ü’. The Tenyidie alphabet has 25 letters, 6 vowels, and 19 consonants. The vowels are: \textit{a,e,i,o,u,ü} and the Consonants are: \textit{b,c,d,f,g,h,j,k,l,m,n,p,r,s,t,v,w,y,z}. An example sentence in Tenyidie is given below:
\begin{center} {\it ‘N shürho kimhie ba ro.}’\\ (How is your health?)
\end{center}

\subsection{Syllable Structure, Formation and Consonant Clusters}
Tenyidie exhibits \textit{monosyllables}~\cite{4}, \textit{disyllables}~\cite{6}, and \textit{sequisyllables}~\cite{8}. Any vowel in Tenyidie can function as the peak of a syllable. In the case of dipthongs the second member being more sonorant functions as the peak, and the first one functions as a glide only. All consonants can occur initially. This means any syllable can be of the form CV~\cite{4}. Kuolie~\cite{4} claims that no consonant can occur as the coda which makes Tenyidie an open syllable language. Kevichüsa~\cite{7} and Ezung~\cite{9,10} however, documents a rhotic consonant as a coda in Tenyidie.

Kuolie~\cite{4} opines following syllable types as the most commonly found. A syllabic peak is also accompanied by one tone.\\
1. One vowel: V\\
2. One consonant followed by a vowel: CV\\
3. An initial consonant cluster followed by one vowel: CCV

Benedict~\cite{11} mentions that Tibeto-Burman consonant clusters are found only in root initial position. Tenyidie is a typical example of having only one type of root initial clusters, plosive plus trill. Not all plosives can form cluster with a trill. Consonant clusters can be formed with the following plosives only~\cite{4}:\\
i. bilabial voiceless plosive: /p + r/	\\
ii. bilabial voiceless aspirated plosive: /ph + r/	\\
iii. velar voiceless plosive: /k + r/	\\	
iv. velar voiceless aspirated plosive: /kh + r/	

\section{Related Works}
\label{section:related works}
This section highlights some of the research conducted on syllabification for the Tibeto-Burman (TB) languages. Much work has been done on syllabification for resource-rich languages like English, Mandarin, German, French,  and Indian Languages like Hindi, Bengali, Tamil, Punjabi, Odia, etc. For Tibeto-Burman Languages, we can find works for languages like \textit{Manipuri}, \textit{Bodo}, etc.

For the TB languages, Sarma et al.~\cite{12} developed an algorithm for the automatic syllabification of the Bodo language by studying Bodo syllable structure and linguistic rules for syllabification. The algorithm was tested on 5,000 words obtained from a corpus and compared with the same words manually syllabified, achieving an accuracy of 97.05\%. Singh et al.~\cite{13} proposed several data-driven methods for automatically syllabifying words written in the Manipuri language, a scheduled Indian language. First, they proposed a language-independent rule-based approach formulated using entropy-based phonotactic segmentation. Second, they projected the syllabification problem as a sequence labeling problem and investigated its effect using various sequence labeling approaches. Third, they combined the effect of sequence labeling and rule-based methods and investigated the performance of the hybrid approach. From various experimental observations, it was observed that the proposed methods outperform the baseline rule-based method. The entropy-based phonotactic segmentation provides a word accuracy of 96\%, CRF (sequence labeling approach) provides 97\%, and the hybrid approach provides 98\% word accuracy. In another work for the Manipuri Language, Devi et al.~\cite{14} proposed an efficient algorithm for automatic syllabification of the Manipuri Language based on syllable rule structure. The algorithm is evaluated on Manipuri words obtained from different sources like textbooks, newspapers, etc. The algorithm's result achieved 99.8\% accuracy as compared to manual syllabification.

\section{Corpus Generation}
\label{section: corpus_generation}
\begin{figure}[t!]
\begin{center}
	\[\includegraphics[scale=0.4]{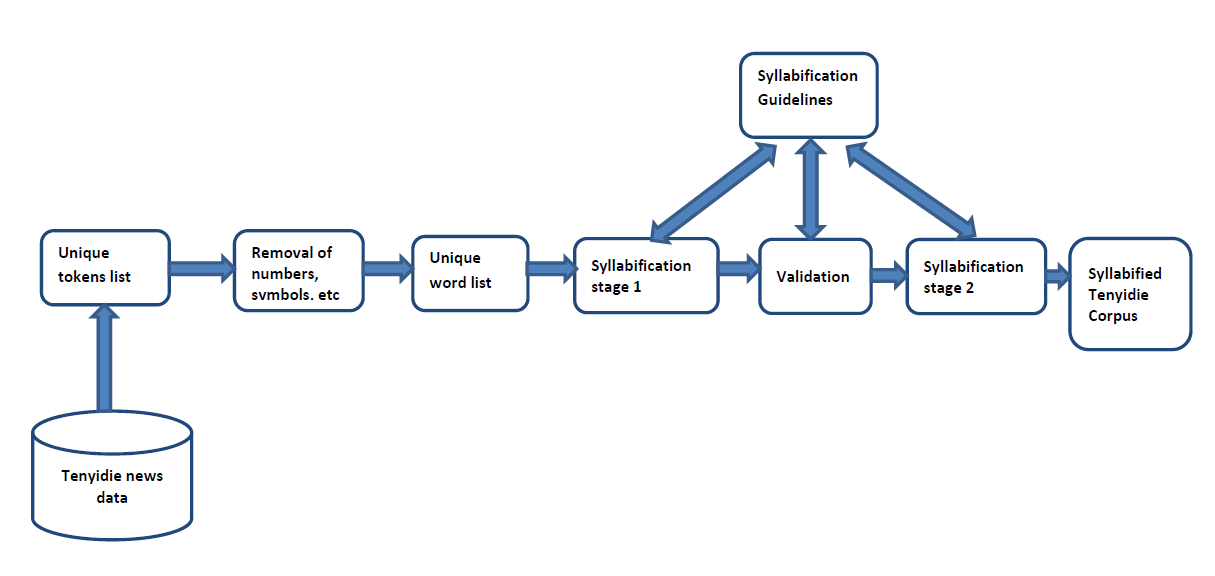} \]
	\caption{Tenyidie manual syllabification methodology}
 \label{fig:methodology}
\end{center}
\end{figure}

To apply any learning model, we need to create a training dataset (annotated data) to train the learning model. This section describes the manual annotated corpus creation process and the statistical analysis of the Tenyidie syllables appearing in our annotated dataset.
The corpus creation process starts with data acquisition in which we collected a total of 16,022 unique words from news data (Tenyidie news daily \textit{"CAPI"}). From this unique words list, we removed the non-Tenyidie words and other tokens (numbers, symbols, etc.), and the remaining 10,120 words are used for our syllabification dataset annotation. 
The unstemmed dataset statistics are given in Table~\ref{table:raw_dataset_statistics}, with a minimum word length of 1, average word length of 8.58, and maximum word length of 20. For the manual annotation process, in the first stage, we start the manual syllabification by the first annotator (Annotator 1) by following the syllabification guidelines given by the language expert. In the second stage, the second annotator (Annotator 2), after consultation with the language expert and Annotator 1, then manually corrects the first annotated dataset. The manual syllabification process is depicted in Fig.~\ref{fig:methodology}.

In the final syllabified corpus, we observed an annotation correction (change in annotated dataset frequency) of 1,104 words between the first and the second annotated dataset, which constitutes 10.9\% of the total annotated dataset. The definite marker \textit{'-u'}, exclusive marker \textit{'-e'}, plural marker \textit{'-ko'}, and nominal base \textit{'-mia'}, wherever the hyphen appears, is attached to the syllable since it is a constituent component of the syllable to form the markers in Tenyidie. For eg, kekuo-u (ke kuo \textit{-u}), mechüu-e (me chü u \textit{-e}), kilonser-ko (ki lon ser \textit{-ko}), bodo-mia (bo do \textit{-mia}). Note that the syllables have been separated using white space. A few examples of common mistakes during manual syllabification are:\\
i. single vowel syllable appearing at the beginning of the word. eg. ivor(i vor).\\
ii. single vowel syllable appearing at the end of the word. eg. tuoi (tuo i).\\
iii. single vowel syllable appearing in the middle of the word. eg. khieümhou (khie ü mhou).\\
iv. words beginning with nominalizer \textit{'ke-'}. eg. kenyü (ke nyü).

\begin{table}[t!]
\centering
\caption{Tenyidie unstemmed dataset statistics}
{\begin{tabular}{@{}lc@{}} \hline
\textbf{Particulars} & \textbf{no. of characters} \\
\hline
max. word length & 20 \\ 
min. word length & 1 \\ 
average word length & 8.58 \\\hline
\end{tabular}}
\label{table:raw_dataset_statistics}
\end{table}

\begin{figure}[t!] 
	\centering
	\[\includegraphics[scale=0.6]{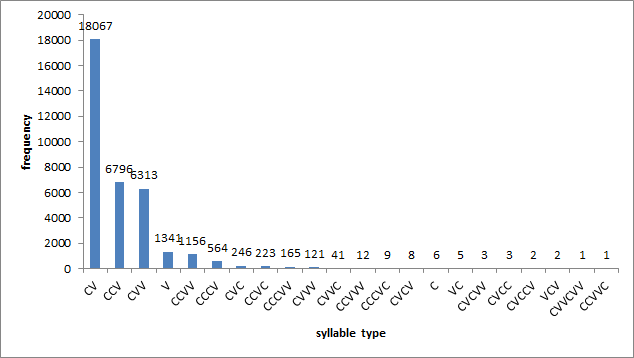} \]
	\caption{Tenyidie syllabified corpus syllable types vs. frequency}
 \label{fig:syllable_vs_frequency}
\end{figure}
\begin{figure}[t!] 
	\centering
	\[\includegraphics[scale=0.6]{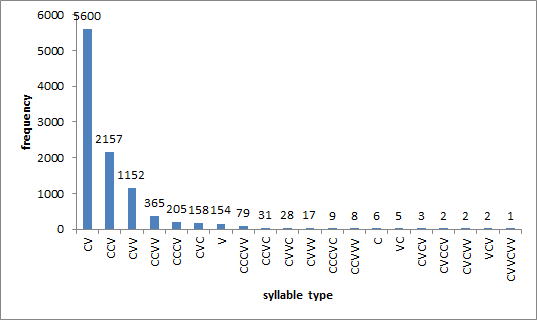} \]
	\caption{Syllable distribution for beginning positioned syllables}
\label{table:syllable_pos_beg}
\end{figure}
\begin{figure}[t!] 
	\centering
	\[\includegraphics[scale=0.65]{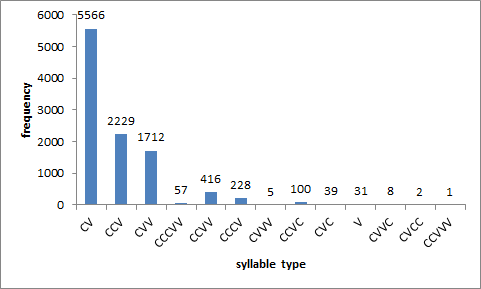} \]
	\caption{Syllable distribution for middle positioned syllables}
\label{table:syllable_pos_middle}
\end{figure}
\begin{figure}[t!] 
	\centering
	\[\includegraphics[scale=0.65]{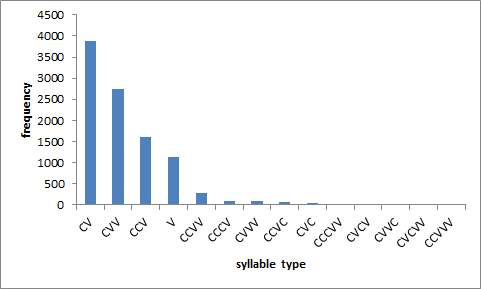} \]
	\caption{Syllable distribution for end positioned syllables}
\label{table:syllable_pos_end}
\end{figure}

Here, we provide statistics of the syllable types available in our manual syllabified corpus. Types of syllables and counts from corpus creation are shown in Fig.~\ref{fig:syllable_vs_frequency}. Apart from the common patterns C, CV, and CVV, we identified the other possible syllable types in Tenyidie. The top 3 (three) syllables are CV, CCV, and CVV. 
We also provide the syllable type frequency with their positional (i.e., beginning, middle, and end) distribution in Fig.~\ref{table:syllable_pos_beg},  Fig.~\ref{table:syllable_pos_middle}, and Fig.~\ref{table:syllable_pos_end} respectively. We observe the top 3 (three) syllable types in the beginning position to be CV, CCV, and CVC; in the middle position to be CV, CCV, and CVV; and in the ending position to be CV, CVV, and CCV. The top 10 syllables with their corresponding frequencies are: \textit{ke (5625), lie (1639), ta (1372), shü (1269), pe (1039), tuo (994), cü (908), u (862), ko (835)} and \textit{ya (784)}. The detailed list of the top 50 syllables can be found in Table~\ref{table: top_50_syllables} in Appendix~\ref{appendix A}.

\begin{figure*}[t!] 
	\centering
	\[\includegraphics[scale=0.12]{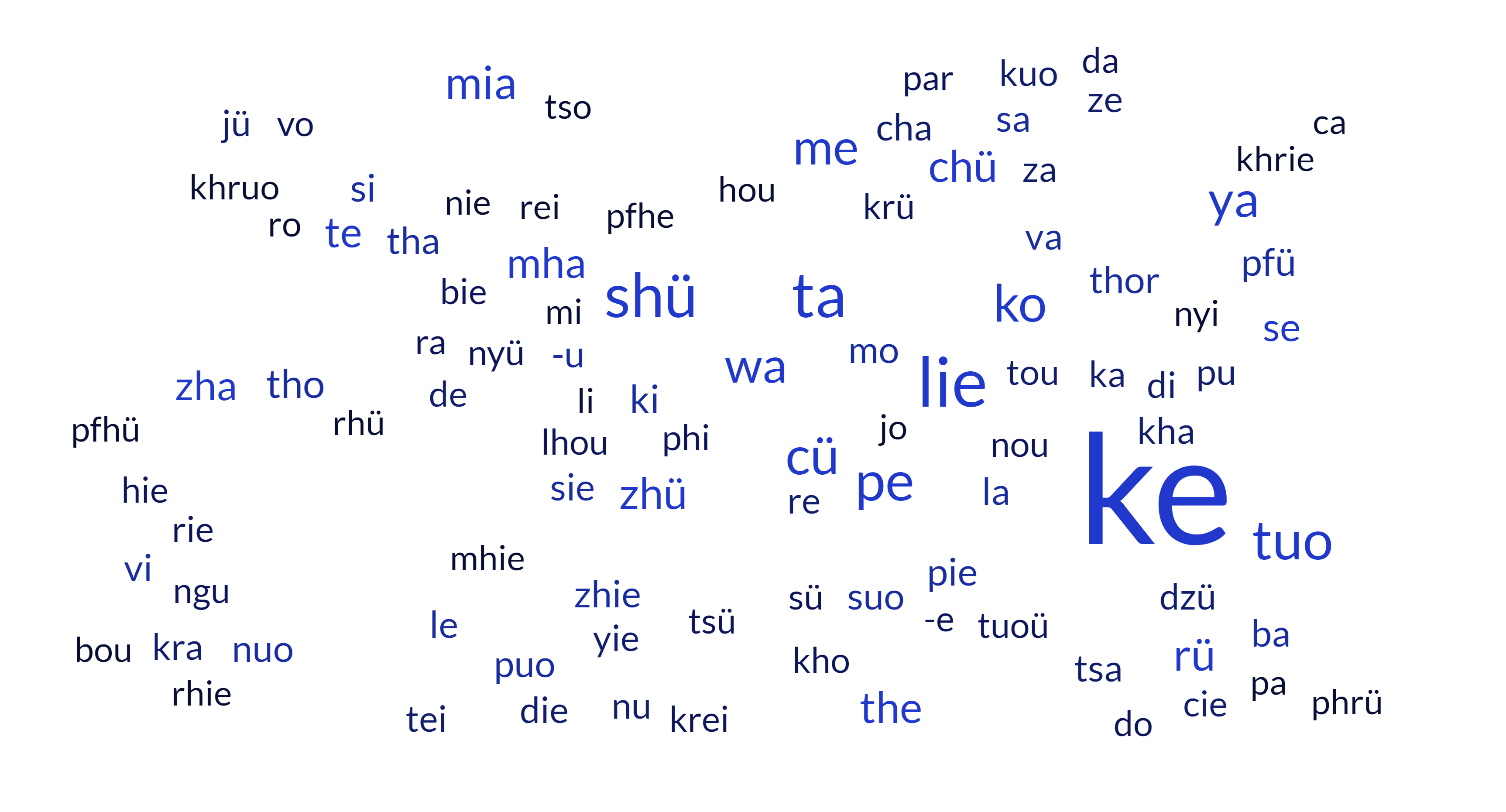} \]
	\caption{Tenyidie syllables distribution word cloud}
\label{table:syllable_word_cloud}
\end{figure*}

\section{Deep Learning experimental setup}
\label{section: learning_model}
The syllabification task is a sequence-to-sequence labeling task in which the input sequence is the word to be syllabified, and the output sequence gives the syllabified sequence of the characters in the input word. To apply the deep learning techniques, we label the syllabified dataset so that the beginning and the end of every syllable in the word can be identified. For example, the word "t e n y i d i e" (te+nyi+die) is labeled as "S C S C C S C C", where the label "S" indicates the starting of the syllable and "C" indicates the continuation of the syllable. Hence, the labeled dataset can be represented as "t/S e/C n/S y i/C d/S i/C e/C". For the experiments, we employed the annotated syllabification dataset with 80, 10, and 10 percent words for training, validation, and testing, respectively, as shown in Table~\ref{table:train_test_dataset}.

\begin{table}[t!]
\centering
\caption{Training, validation, and test dataset}
{\begin{tabular}{@{}lc@{}} \hline
\textbf{Particulars} & \textbf{No. of words} \\
\hline
Total & 10,120 \\ 
Training & 8,096 \\ 
Validation & 1,012 \\ 
Test & 1,012 \\\hline
\end{tabular}}
\label{table:train_test_dataset}
\end{table}

\begin{figure}[t!] 
	\centering
	\[\includegraphics[scale=0.35]{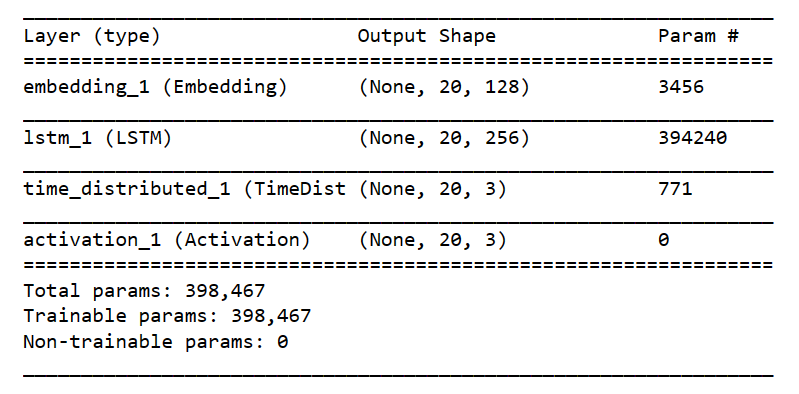} \]
	\caption{LSTM model summary}
    \label{fig:lstm-model}
\end{figure}

\begin{figure}[t!] 
	\centering
	\[\includegraphics[scale=0.35]{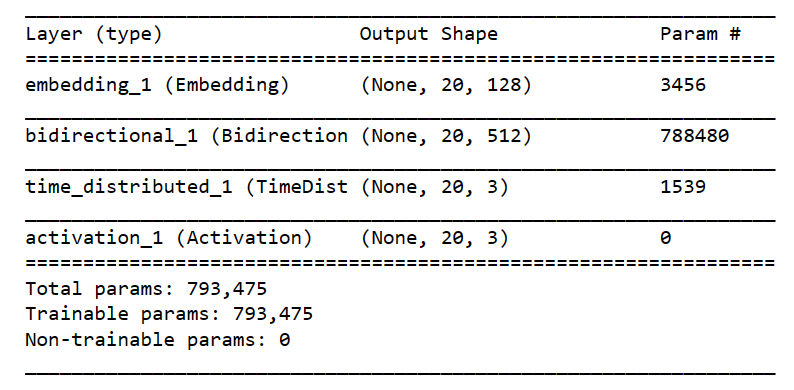} \]
	\caption{BLSTM model summary}
    \label{fig:blstm-model}
\end{figure}

\begin{figure}[t!]  
	\centering
	\[\includegraphics[scale=0.35]{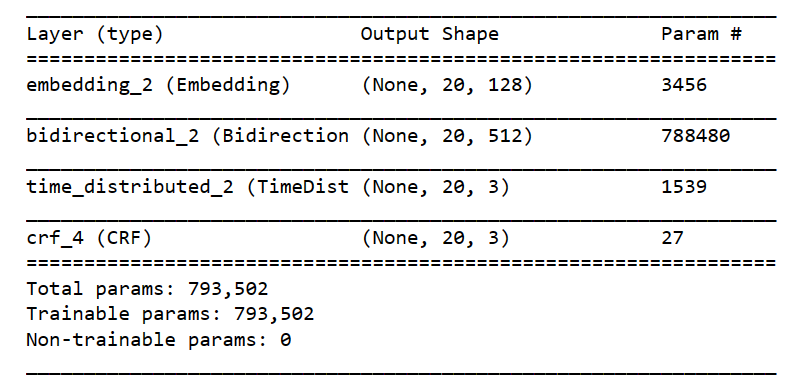} \]
    \caption{BLSTM+CRF model summary}
    \label{fig:blstm+crf-model}
\end{figure}

In our experimental setup, we applied LSTM (Long short-term Memory), BLSTM (Bidirectional Long short-term Memory), BLSTM+CRF (Conditional Random Fields), and Encoder-decoder (Neural Machine Translation) models since these models have proven to give good results in sequence labeling tasks. We implemented LSTM, BLSTM, and BLSTM+CRF models using the Keras (https://keras.io) framework. We show the Keras model summary details of LSTM, BLSTM, and BLSTM+CRF in Fig.~\ref{fig:lstm-model}, Fig.~\ref{fig:blstm-model} and Fig.~\ref{fig:blstm+crf-model} respectively. To initialize the words, we employed a word embedding dimension of 128. The parameters are optimized using the Adam~\cite{15} optimizer, with a batch size of 128. The models underwent training for 40 epochs, with a learning rate of 0.001. The LSTM framework has a trainable parameter of 398,467 parameters, BLSTM with 793,475 trainable parameters, and BLSTM+CRF with trainable 793,502 parameters. The time\_distributed\ layer is used in Keras when using LSTM models to capture the sequence information in a time-recorded manner. 

\begin{figure}[t!] 
	\centering
	\[\includegraphics[scale=0.35]{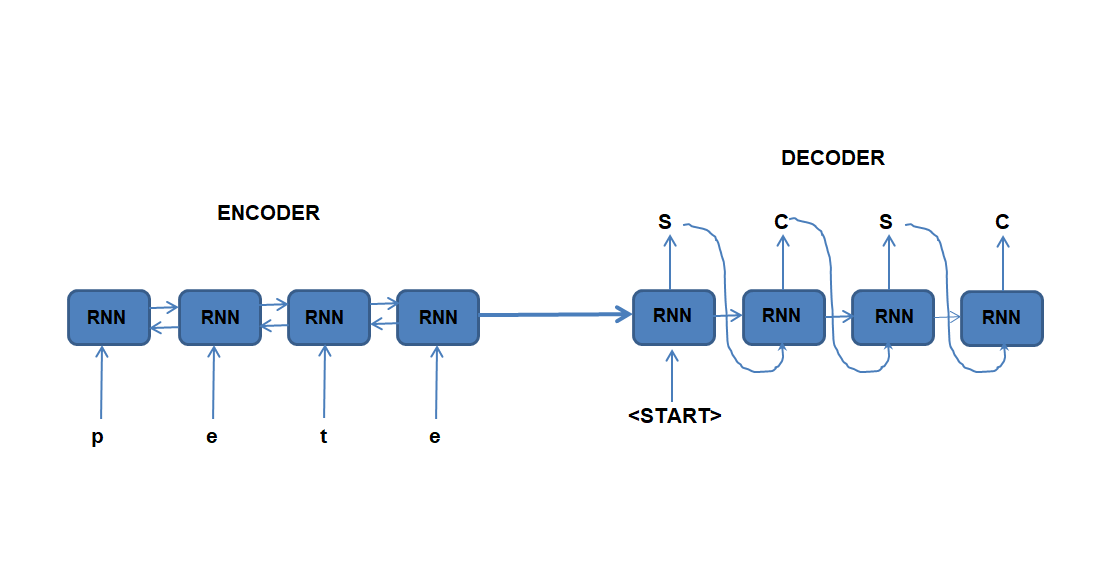} \]
	\caption{Encoder-decoder model overview}
    \label{fig:encoder-decoder-model}
\end{figure}

For the Encoder-decoder NMT model, we used the implementation available in tensorflow.org\footnote{https://www.tensorflow.org/text/tutorials/nmt\_with\_attention}. Fig.~\ref{fig:encoder-decoder-model} shows an overview of the Encoder-decoder model with an illustration of the input and the output sequence. The encoder appears on the left side of the architecture, i.e., the input/source sequence ("p e t e"), and the decoder appears on the right side of the architecture, i.e., the target sequence ("S C S C"). At each time step, the decoder's output is combined with the encoder's output to predict the next character in the target sequence. The goal of the encoder is to process the context sequence into a sequence of vectors for which we use a \textit{bidirectional RNN} to take care of this processing step. The attention layer~\cite{16} then lets the decoder access the information extracted by the encoder. It allows the model to focus on specific sections of the input while executing by dynamically assigning weights to different elements in the input, indicating their relative importance or relevance. It computes a vector from the entire context sequence and adds that to the decoder's output. The decoder then generates predictions for the next token at each location in the target sequence by using a unidirectional RNN to process the target sequence. To perform the experiments, we use batch\_size of 16, embedding\_dim for the input to be 128, RNN units to be 512, and trained over 40 epochs. Note that the other parameters remain the same.

\section{Experimental Results \& Discussions}
\label{section: experimental_results}
This section reports the syllabification results for our deep learning experiments. The accuracy of the syllabification is evaluated at the word level and is calculated as:- \\

\textbf{Accuracy (\%) = \textit{no of words syllabified correctly/Total no. of words}} x 100\%.\\

\begin{figure}[t!] 
\centering
    \centering\includegraphics[width=6cm]{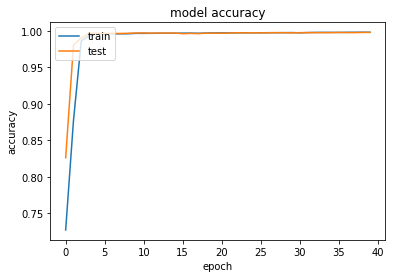}
    \caption{LSTM model Accuracy}
    \label{fig:lstm_acc}
\end{figure}

\begin{figure}[t!]   
    \centering\includegraphics[width=6cm]{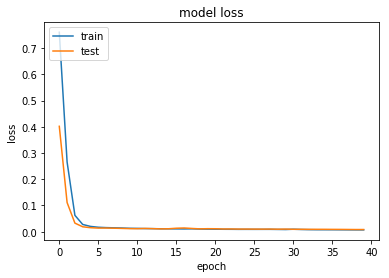}
    \caption{LSTM model Loss}
    \label{fig:lstm_loss}
\end{figure}

\begin{figure}[t!]
    \centering\includegraphics[width=6cm]{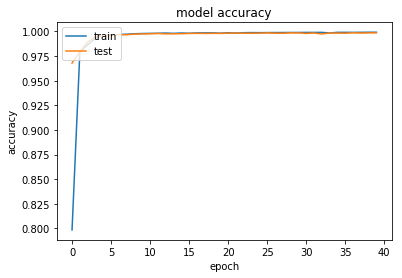}
    \caption{BLSTM model Accuracy}
    \label{fig:blstm_acc}
\end{figure}

\begin{figure}[t!]  
    \centering\includegraphics[width=6cm]{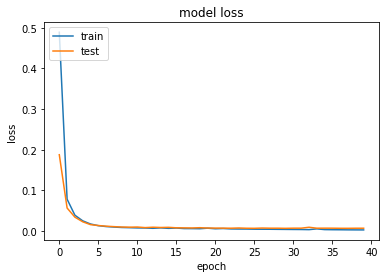}
    \caption{BLSTM model Loss}
    \label{fig:blstm_loss}
\end{figure}

\begin{figure}[t!]
    \centering\includegraphics[width=6cm]{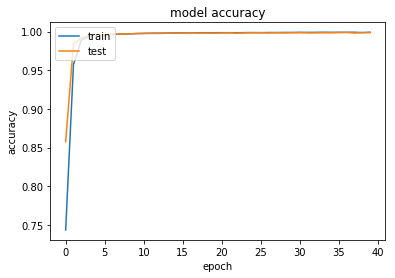}
    \caption{BLSTM+CRF model Accuracy}
    \label{fig:blstm+crf_acc}
\end{figure}

\begin{figure}[t!]
    \centering\includegraphics[width=6cm]{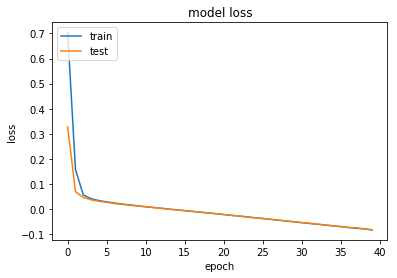}
    \caption{BLSTM+CRF model Loss}
    \label{fig:blstm+crf_loss}
\end{figure}

\begin{figure}[t!]
    \centering\includegraphics[width=6cm]{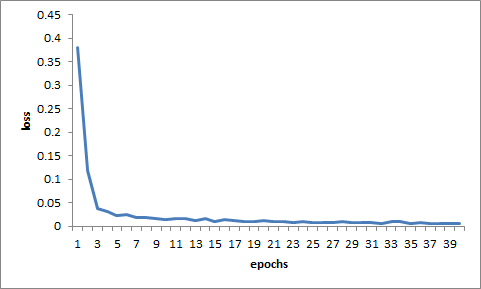}
    \caption{Encoder-decoder  model Loss}
    \label{fig:encoder_gru_loss}
\end{figure}

\begin{figure}[t!]
    \centering
	\[\includegraphics[scale=0.25]{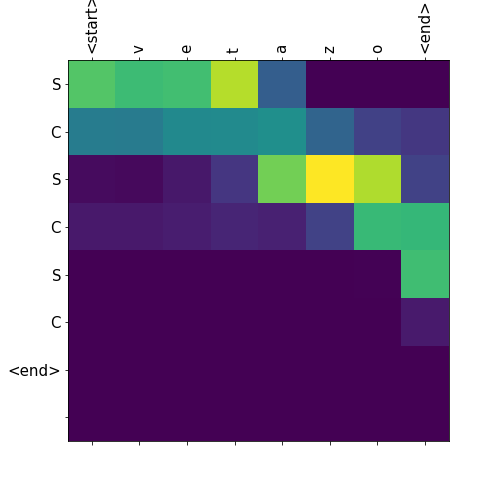} \]
	\caption{Example 1 attention plot (correctly syllabified)}
	\label{fig:attention-plot-1}
\end{figure}

The test data for the syllabification comprises of 1,012 words, which is 10\% of the total annotated dataset. The syllabification results are shown in Table~\ref{table: result_evaluation}. We observed a syllabification accuracy of 97.04\% for LSTM, 99.21\% for BLSTM, 99.01\% for BLSTM+CRF, and 94.27\% for Encoder-decoder with attention. The highest accuracy is obtained by BLSTM, with a syllabification accuracy of 99.21\%. The model accuracies are shown in Fig.~\ref{fig:lstm_acc}, Fig.~\ref{fig:blstm_acc}, Fig.~\ref{fig:blstm+crf_acc} and model losses are shown in Fig.~\ref{fig:lstm_loss}, Fig.~\ref{fig:blstm_loss}, Fig.~\ref{fig:blstm+crf_loss}, and Fig.~\ref{fig:encoder_gru_loss}. We can observe that the model accuracies and the loss attain good results within the first five (5) epochs. Fig.\ref{fig:attention-plot-1}  show the attention heatmap of an example Tenyidie word which is correctly syllabified. By referring to the heatmaps, we can get information on which characters are given more importance/relevance in determining the syllable breakup. The attention heatmaps can be studied in the future while building rules for developing rule-based syllabification algorithms for Tenyidie. 

In our experiments, we have tested on 10\% of our dataset due to the limited size of our annotated dataset for performing other evaluation techniques such as cross-validation, etc. It is also observed that Encoder-decoder models could not perform better than BLSTM since these models perform well on larger datasets.

In Table~\ref{table:error_analysis}, we report the error analysis of the BLSTM experiment since it achieved the best performance. In the table, we show the word with the syllable formation, the CV syllable pattern, the actual tagging sequence, and the predicted tagging sequence. It is observed that most of the errors occur with different markers such as 'u' (definite marker) and 'ü'. For eg, the word 'shesou' should be tagged as \textit{she+so+u} but tagged as \textit{she+sou}. Table~\ref{table:error_comparison_analysis} provides the comparison of error analysis for BLSTM output with other models. The output for the other models that predicted correctly for the wrong prediction of the BLSTM are indicated in bold. We observe four (4) correct predictions for LSTM, five (5) correct predictions for BLSTM+CRF, and one (1) correct prediction for Encoder-decoder. In Table~\ref{table:incorrect_output_comparison}, we provide a comparison of the incorrect syllabified words by different models. 

\begin{table}[t!]
\caption{Evaluation of deep learning models}
{\begin{tabular}{@{}lcc@{}} \hline
\textbf{DL technique} & \textbf{\begin{tabular}[c]{@{}l@{}}no. of words\\ syllabified correctly\end{tabular}} & \textbf{Accuracy (\%)} \\\hline
{LSTM} & 982/1012 & 97.04\% \\
\textbf{BLSTM} & \textbf{1004/1012} & \textbf{99.21}\% \\
{BLSTM+CRF} & 1002/1012 & 99.01\% \\
{\begin{tabular}[c]{@{}l@{}}Encoder-decoder (NMT) with attention \end{tabular}} & 954/1012 & 94.27\% \\ 
\hline
\end{tabular}}
\label{table: result_evaluation}
\end{table}

\hspace*{-15cm}
\begin{table}[t!]
\caption{Error analysis for the BLSTM output}
	\small
{\begin{tabular}{@{}llll@{}} \hline
\textbf{word} & \textbf{syllable pattern} & \textbf{actual} & \textbf{predict} \\\hline
chüümo-u (chü+ü+mo+-u) & CCV V CV V & SCCSSCS & SCCSCCS\\
tseiü (tsei+ü) & CCVV V & SCCCS & SCCCC\\
cieürielieketuo (cieü+rie+lie+ke+tuo) & CVVV CVV CVV CVV & SCCCSCCSCCSCSCC & SCCSCCCSCCSCSCC\\
rüünuo (rü+ü+nuo) & CV V CVV & SCSSCC & SCCSCC\\
thepfhetheü (the+pfhe+theü) & CCV CCCV CCVV & SCCSCCCSCCC & SCCSCCCSCCS\\
mezhükieo (me+zhü+ki+e+o) & CV CCV CV V V & SCSCCSCSS & SCSCCSCSC\\
shesou (she+so+u) & CCV CV V & SCCSCS & SCCSCC\\
pedashütaü (pe+da+shü+taü) & CV CV CCV CVV & SCSCSCCSCC & SCSCSCCSCS
 \\\hline
\end{tabular}}
\label{table:error_analysis}
\end{table}

\begin{sidewaystable}
\centering
	\caption{Comparison of BLSTM error analysis with other models (correctly labeled indicated in bold)}
	\scriptsize
{\begin{tabular}{@{}llllll@{}} \hline
\textbf{word} & \textbf{actual}& \textbf{BLSTM output} & \textbf{LSTM output} & \textbf{BLSTM+CRF output} & \textbf{Encoder-decoder output} \\\hline
chüümo-u (chü+ü+mo+-u) &	SCCSSCS &	SCCSCCS &	\textbf{SCCSSCS} &	\textbf{SCCSSCS} &	SCCCSCS\\
tseiü (tsei+ü) &	SCCCS &	SCCCC &	\textbf{SCCCS} &	\textbf{SCCCS} &	SCCCC\\
cieürielieketuo (cieü+rie+lie+ke+tuo) &	SCCCSCCSCCSCSCC &	SCCSCCCSCCSCSCC &	SCCSCCCSCCSCSCC &	SCCSSCCSCCSCSCC &	\textbf{SCCCSCCSCCSCSCC}\\
rüünuo (rü+ü+nuo) &	SCSSCC &	SCCSCC &	SCCSCC &	SCCSCC &	SCCSCC\\
thepfhetheü (the+pfhe+theü)&	SCCSCCCSCCC&	SCCSCCCSCCS&	\textbf{SCCSCCCSCCC}&	\textbf{SCCSCCCSCCC}&	SCCSCCCSCCS\\
mezhükieo (me+zhü+ki+e+o)&	SCSCCSCSS&	SCSCCSCSC&	SCSCCSCCC&	\textbf{SCSCCSCSS}&	SCSCCSCCC\\
shesou (she+so+u)&	SCCSCS&	SCCSCC&	SCCSCC&	\textbf{SCCSCS}&	SCCSCC\\
pedashütaü (pe+da+shü+taü)&	SCSCSCCSCC&	SCSCSCCSCS&	\textbf{SCSCSCCSCC}&	SCSCSCCSCS&	SCSCSCCSCS\\\hline
\end{tabular}}
\label{table:error_comparison_analysis}
\end{sidewaystable}

\begin{table}[t!]
\caption{Table showing incorrect output comparison by different models}
\scriptsize
{\begin{tabular}{llll}\toprule
\textbf{LSTM} &
  \textbf{BLSTM} &
  \textbf{BLSTM+CRF} &
  \textbf{NMT WITH ATTENTION} \\\midrule
\begin{tabular}[c]{@{}l@{}}ndunu (n du nu)\\ vohutazhie (vo hu ta zhie)\\ miapuoe (mia puo e)\\ kerlieya (ker lie ya)\\ shier (shier)\\ cieürielieketuo (cieü rie lie ke tuo)\\ rüünuo (rü ü nuo)\\ lavorketa (la vor ke ta)\\ kesewa (ke se wa)\\ merieyhara (me rie yha ra)\\ phirlieketa (phir lie ke ta)\\ ani (a ni)\\ khüsoh (khü soh)\\ kepartsa (ke par tsa)\\ tirza (tir za)\\ galil (ga lil)\\ dikon (di kon)\\ ndunhiewe (n du nhie we)\\ rütsorümeiko (rü tso rü mei ko)\\ vorü (vo rü)\\ kerlieketa (ker lie ke ta)\\ chülieketuoe (chü lie ke tuo e)\\ mezhükieo (me zhü ki e o)\\ lertaketuo (ler ta ke tuo)\\ shijoh (shi joh)\\ chotisuh (cho ti suh)\\ kedahlo (ke dah lo)\\ zecülietaü (ze cü lie ta ü)\\ shesou (she so u)\\ chakan (cha kan)\end{tabular} &
  \begin{tabular}[c]{@{}l@{}}chüümo-u \\(chü ü mo -u)\\ tseiü \\(tsei ü)\\ cieürielieketuo \\(cieü rie lie ke tuo)\\ rüünuo \\(rü ü nuo)\\ thepfhetheü \\(the pfhe theü)\\ mezhükieo \\(me zhü ki e o)\\ shesou \\(she so u)\\ pedashütaü \\(pe da shü taü)\end{tabular} &
  \begin{tabular}[c]{@{}l@{}}jüü \\(jüü)\\ chüüphou \\(chü ü phou)\\ miapuoe \\(mia puo e)\\ cieürielieketuo \\(cieü rie lie ke tuo)\\ rüünuo \\(rü ü nuo)\\ merieyhara \\(me rie yha ra)\\ tirza \\(tir za)\\ sikecüumonyü \\(si ke cü u mo nyü)\\ avolü \\(a vo lü)\\ pedashütaü \\(pe da shü taü)\end{tabular} &
  \begin{tabular}[c]{@{}l@{}}siliekecü (si lie ke cü)\\ kerheiro-u (ke rhei ro -u)\\ chüümo-u (chü ü mo -u)\\ vohutazhie (vo hu ta zhie)\\ leshümhasi-e (le shü mha si -e)\\ keyiethortazhie (ke yie thor ta zhie)\\ pekaperuowazhü (pe ka pe ruo wa zhü)\\ kinyi-e (ki nyi -e)\\ tseiü (tsei ü)\\ dzüchakou (dzü cha ko u)\\ ketsokemiekecüu (ke tso ke mie ke cü u)\\ kevi-e (ke vi -e)\\ kinyibo-u (ki nyi bo -u)\\ dzüu (dzü u)\\ rüünuo (rü ü nuo)\\ pekralieyakezhamia (pe kra lie ya ke zha mia)\\ kesewa (ke se wa)\\ merieyhara (me rie yha ra)\\ ani (a ni)\\ tsielhoutuo-e (tsie lhou tuo -e)\\ tirza (tir za)\\ bakbukia (bak bu kia)\\ thelhibo-u (the lhi bo -u)\\ kemesawaketau (ke me sa wa ke ta u)\\ nanyü-rüla (na nyü - rü la)\\ pfüe (pfü e)\\ setuoü (se tuoü)\\ dikonpfü (di kon pfü)\\ thepfhetheü (the pfhe theü)\\ doulie (dou lie)\\ lhitho-ue (lhi tho -u e)\\ vorü (vo rü)\\ biepesuowaketa (bie pe suo wa ke ta)\\ zhimomi-e (zhi mo mi -e)\\ mezhükieo (me zhü ki e o)\\ pebaketuou (pe ba ke tuo u)\\ se-dia (se - dia)\\ keyieshüyakezhau (ke yie shü ya ke zha u)\\ chishi-e (chi shi -e)\\ vikhoo (vi kho o)\\ tsiarie-u (tsia rie -u)\\ pekhro-u (pe khro -u)\\ pevowataü (pe vo wa ta ü)\\ diebo-u (die bo -u)\\ vechita-e (ve chi ta -e)\\ zuo (zu o)\\ chotisuh (cho ti suh)\\ petsashükewako (pe tsa shü ke wa ko)\\ themo-u (the mo -u)\\ zecülietaü (ze cü lie ta ü)\\ shesou (she so u)\\ abeio (a beio)\\ pedashütaü (pe da shü taü)\\ pfhekholieketau (pfhe kho lie ke ta u)\\ thenumia-e (the nu mia -e)\\ thedzekethu-ue (the dze ke thu -u e)\\ pethashüyakezha (pe tha shü ya ke zha)\\ suothorzhüketa (suo thor zhü ke ta)\end{tabular}\\\hline
\end{tabular}}
\label{table:incorrect_output_comparison}
\end{table}

\section{Conclusion \& Future works}
\label{section: conclusion}
In this paper, we worked on syllabification for the Tenyidie Language. The main contribution of this work is the creation of the manually annotated syllabification dataset of 10,120 words. We have applied various deep learning architectures such as LSTM, BLSTM, BLSTM+CRF, and Encoder-decoder model on our created dataset. On a test dataset of 1,012 words (i.e., 10\% of the total syllabified dataset), we achieved the highest accuracy of 99.21\% with the BLSTM model. In the future, we will use the syllabifier in various syllable-based NLP tasks, such as studying morphological characteristics, part-of-speech tagging, machine translation, etc. It is also intended to annotate a larger dataset and see the performance of the Encoder-decoder model since these models perform well on larger datasets.

\section*{Acknowledgements}
We are thankful to Dr. Keneichanuo Mepfhüo for her immense help while creating the Tenyidie syllabified corpus.

\bibliographystyle{ws-ijalp}

\newpage
\appendix
\section{Syllables and frequency (Top 50)}
\label{appendix A}
\begin{table}[t!]
\scriptsize
\centering
\caption{Syllables and frequency (Top 50)}
\begin{tabular}{|c|c|c|}
\hline
\textbf{Sl. No} & \textbf{syllable} & \textbf{syllable frequency} \\ \hline
1 & ke & 5625\\
2 & lie & 1639\\
3 & ta & 1372\\
4 & shü & 1269\\
5 & pe & 1039\\
6 & tuo & 994\\
7 & cü & 908\\
8 & u & 862\\
9 & ko & 835\\
10 & ya & 784\\
11 & me & 737\\
12 & wa & 737\\
13 & mia & 556\\
14 & rü & 512\\
15 & chü & 491\\
16 & zhü & 467\\
17 & the & 459\\
18 & mha & 416\\
19 & te & 395\\
20 & zha & 342\\
21 & se & 287\\
22 & tho & 280\\
23 & ki & 243\\
24 & pfü & 230\\
25 & pie & 220\\
26 & si & 197\\
27 & vi & 192\\
28 & le & 185\\
29 & la & 185\\
30 & thor & 185\\
31 & tha & 179\\
32 & zhie & 178\\
33 & va & 174\\
34 & ba & 172\\
35 & sie & 168\\
36 & nuo & 167\\
37 & suo & 167\\
38 & puo & 166\\
39 & mo & 165\\
40 & -u & 163\\
41 & sa & 161\\
42 & cha & 159\\
43 & tsa & 157\\
44 & kha & 149\\
45 & pu & 142\\
46 & die & 139\\
47 & ze & 139\\
48 & di & 129\\
49 & kra & 129\\
50 & jü & 127\\\hline
\end{tabular}
\label{table: top_50_syllables}
\end{table}

%
%

\end{document}